\documentclass[letterpaper]{article}

\usepackage{natbib,alifeconf}  

\usepackage{lipsum}
\usepackage{graphicx}
\usepackage[utf8]{inputenc}

\usepackage{amsmath}
\usepackage{url}

\usepackage[font=footnotesize]{subfig}

\title{
    Real-time Evolution of Multicellularity with Artificial Gene Regulation
}
\author{Dylan Cope \\
\mbox{}\\
King's College London \\
\url{dylan.cope@kcl.ac.uk}
} 

\usepackage{fancyhdr}
\fancypagestyle{specialfooter}{%
  \fancyhf{}
  
  \fancyfoot[R]{Accepted for publication in the proceedings of the 2023 Conference on Artificial Life}
}
\begin{document}
\maketitle
\thispagestyle{specialfooter}
\begin{abstract}
This paper presents a real-time simulation involving ``protozoan-like'' cells that evolve by natural selection in a physical 2D ecosystem. Selection pressure is exerted via the requirements to collect mass and energy from the surroundings in order to reproduce by cell-division. Cells do not have fixed morphologies from birth; they can use their resources in \emph{construction projects} that produce functional nodes on their surfaces such as photoreceptors for light sensitivity or flagella for motility. Importantly, these nodes act as modular components that connect to the cell's control system via IO channels, meaning that the evolutionary process can replace one function with another while utilising pre-developed control pathways on the other side of the channel. A notable type of node function is the adhesion receptors that allow cells to bind together into multicellular structures in which individuals can share resource and signal to one another. The control system itself is modelled as an artificial neural network that doubles as a \emph{gene regulatory network}, thereby permitting the co-evolution of form and function in a single data structure and allowing cell specialisation within multicellular groups.
\end{abstract}

\section{Introduction}

Applying the principles of evolution by natural selection to developing complex systems within simulations is as old as the field of computer science itself, with Alan Turing being an early proponent of the idea. But since then, much has changed in our understanding of evolution, notably with the roles that epigenetic and developmental processes play in mediating the expression of genetic information for multicellular life. 

This paper presents a complex bio-inspired simulation involving cell-like entities that move around, gather resources, grow, engage in `construction projects' to develop traits, reproduce by cell-division, and bind together to form multicellular structures. The environment is a 2-dimensional continuous space and cells are rigid discs that can move and collide with obstacles according to Newtonian mechanics. In order to grow in size or construct traits, a cell needs mass and energy. The primary source of this is ``plant cells'' that also grow and divide in the environment.

Traits are derived from the outputs of a recurrent artificial Gene Regulatory Network (GRN), and thus are potentially dynamic and subject to environmental influence or signals from other cells (including the parent). To implement this a system of modular functions that interface with the cell's GRN is introduced that grow at sites on the cell's surface. The following sections outline the background literature on this topic, the mechanics of the simulation, and observations from running the simulation and letting it evolve.

The key contributions of this work are: 1) demonstrating the emergence of multicellular cell specialisation within a real-time evolution simulation, 2) the introduction of a novel method for representing evolvable traits as modular components that develop from a GRN via a \textit{`fuzzy lock-and-key'} algorithm.

\section{Background}

\subsection{Formal Models and Simulations of Evolution}

Computational models of evolution have been studied in a wide variety of forms. They are considered to be an important tool for investigating Darwinian dynamics and the results often surprise researchers \citep{lehman_surprising_2018}. A fundamental hurdle posed by the challenge is the vast amounts of computation and serendipity that real-life evolutionary processes required. The aeons of interactions at the microscopic scale operating across the face of planet are not easily simplified or replicated. So, in order to understand various processes at a higher level we develop abstractions to ground our models upon. The following is a short tour of approaches at various levels-of-abstraction, so as to contextualise the simulation presented in this work.

At the most abstract end, we have models that evolutionary populations as players in a strategic game; this is the field of \textit{Evolutionary Game Theory} \citep{alexander_evolutionary_2021}. 
Sliding down the abstraction ladder, we encounter models that incorporate varying degrees of physical interactions. In cellular automata-based simulations the update rules of the grid cells may be viewed as a basic `physics', or in other simulations entities may interact as soft or rigid bodies in continuous motion under Newtonian mechanics \citep{joachimczak_fine_2014, komosinski_world_2000}. Parallel to this, simulations may also get closer to real processes by means of having survival rates emerge from resource gathering and reproduction \citep{lenski_evolutionary_2003, yedid_microevolution_2001, collins_antfarm_1991}. This is in contrast to explicitly represented fitness functions that condition survival on solving a given engineering problem unrelated to the evolutionary process itself.

At the `lowest level' of this hierarchy is methods that directly model biochemical processes. This area of \textit{in silico genetics} has involved modelling networks of transcription factors \citep{jenkins_new_2009}, gene expression \citep{liao_silico_2004}, the evolution of metabolic chemistry \citep{ullrich_silico_2011}, or molecular evolution \citep{schuster_extended_1993}.

\subsection{Artificial Gene Regulation and Simulated Multicellularity}

Of specific interest to this project are works that involve simulating gene regulatory networks (GRNs) within evolutionary processes. One of the most widely regardly early approaches to this was Kauffman's Random Boolean Networks; graphs in which nodes are related to one another through state update rules expressed as statements of propositional logic \citep{kauffman_homeostasis_1969}. For a full review of artificial GRNs see \cite{cussat-blanc_artificial_2018}. The most similar models to the GRNs in this work are the neural network-based GRNs that resided in each cell of the multicellular animats evolved for locamotion \citep{joachimczak_fine_2014}.

\section{Simulation Mechanics}

This section outlines the core mechanisms underlying the simulation, starting with the basic `biological' systems, physics, and resource flows. We then discuss the modular surface nodes system, and outline how reproduction and the gene regulatory networks are implemented.

\begin{figure*}[!ht]
\centering
\subfloat[Closer view of white protozoa cells feeding via osmosis and phagocytosis. The osmotrophy is demonstrated by trails of absence left behind protozoa. Phagocytosis is visible as plant/meat cells being engulfed.\label{subfig-1:chemical-soln-cells-feeding}]{%
      \includegraphics[width=.48\linewidth]{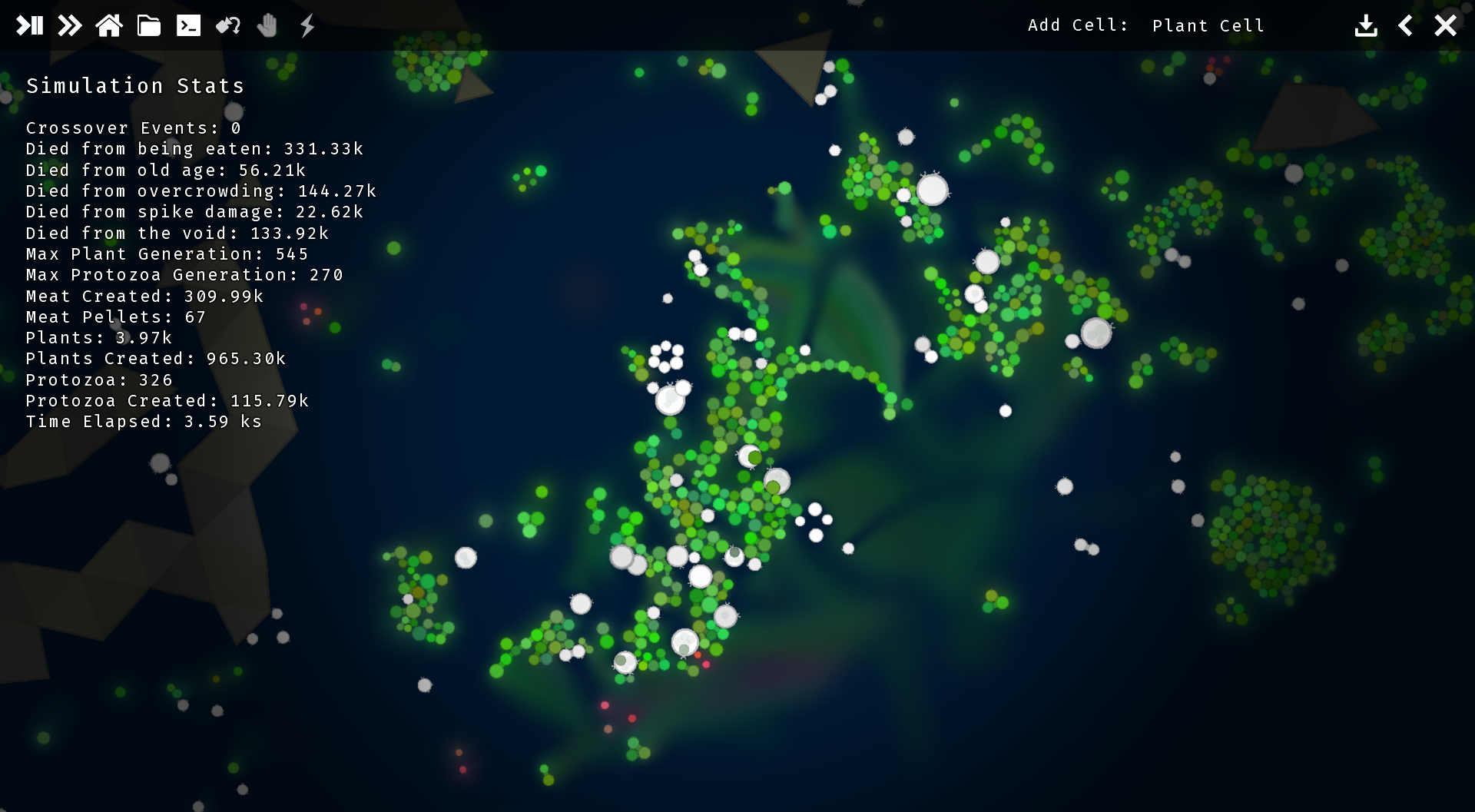}
}
\hspace{.2cm}
\subfloat[A zoomed-out view of the environment. The `rock' structures can be seen, creating variation in the environment. Large masses of green plants are also visible. Individual protozoan cells are not very visible.\label{subfig-2:zoomed-out-view}]{%
      \includegraphics[trim={0 .5cm 0 0},clip,width=.48\linewidth]{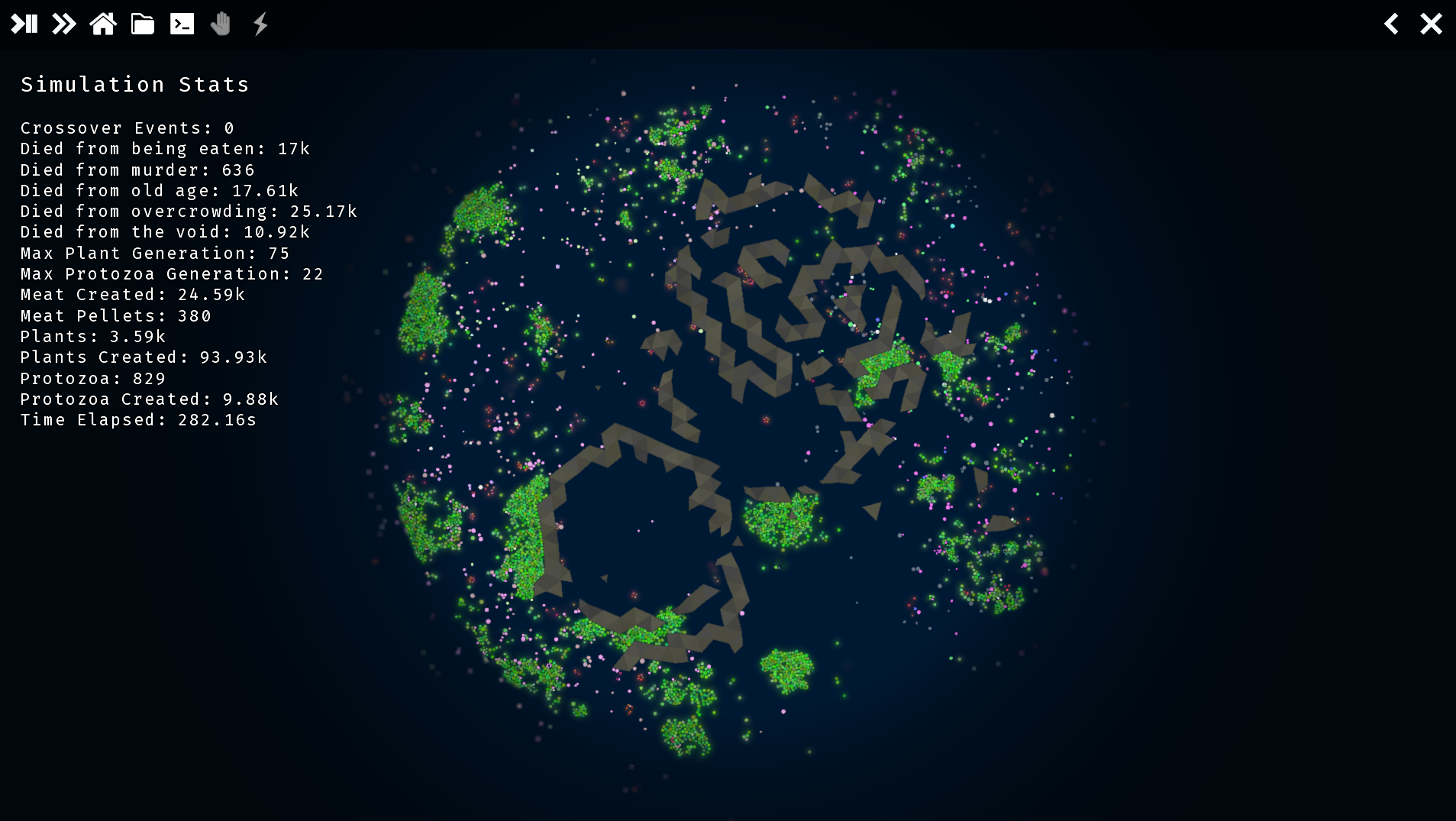}
}
\caption{Zoomed-in and zoomed-out views in the simulation UI.}\label{fig:sim-first-look}
\end{figure*}

\subsection{Physics, Resources, and Cell Basics}

\textbf{Physical Environment.} The simulation consists of various physical bodies that interact with one another and move according to Newtonian mechanics. There are static bodies that are fixed in place and dynamic bodies that move according to impulses and torques applied to them. Joints can stick cells together by restricting their movements to keep the joint anchors (placed on each body) a given distance away. Joint anchors are an important component for allowing multicellular structures to retain their form as they are perturbed as they keep joined cells at roughly consistent relative angles to one another. Movement is linearly dampened to simulate the effect of the cells being light bodies in a viscous fluid. The dampening is reduced for cells in multicellular structures where drag is reduced proportionally to the cosine distance between the velocity vector and the difference in cell positions. Therefore, a group of cells arranged into a line would face more resistance moving in the direction parallel to the line than the direction perpendicular. Stepping the physics simulation is handled by the Box2D library.

The environment is composed of a central circular region containing `rock formations'; groups of contiguous rigid boundaries formed of triangles. These are procedurally generated and add environmental diversity in terms of imposing different modes of movement restrictions on cells. Typically, randomly generated environments have the following characteristics. Some areas are wide open without any boundaries nearby, others have a moderate amount of space enclosed within a mostly-connected ring of rocks, and others are tight spaces consisting of snaking tunnels. Outside of the central area is the `void', an empty region that cells can enter into, but where they begin to lose health proportionally to how far they venture. 

\textbf{Complex Molecules.} Objects in the simulation referred to as `complex molecules' are resources that serve as prerequisites for constructing various traits (see the next section). These are represented by a number in the unit interval called the \emph{molecule's signature}. The simulation only permits a finite number of possible molecules, represented at evenly spaced intervals between 0 and 1. For example, in the runs conducted this was configured to 128 possibilities, so the molecule indexed at $i$ in this set had the signature $i/128$. Complex molecules are produced using mass and energy according to a `production cost` in terms of energy expended per unit mass produced.

The term `complex molecules' is inspired by the \emph{Central Dogma of Molecular Biology}, that states: ``DNA codes for RNA, and RNA codes for proteins''. Complex molecules in the simulation are designed to be analogous to proteins in the following ways; firstly, they are what ultimately implement functions in a cell. Secondly, they are involved in `lock-and-key' mechanisms that act as conditional switches (more on this when we discuss how the construction projects determine cell node functionality in the simulation). Thirdly, the choice of which complex molecules are produced is made by the gene regulatory network.

\textbf{Cell Attributes.} All cells maintain several basic attributes; firstly, a health value. This starts with a value of 1 upon a cell's birth and steadily decreases, until it drops below 0.05 and the cell dies. Mass and energy can be used to repair the cell, but this can only be done when there is an excess as priority is given to growth by default (although mechanisms can evolve to reverse this priority). The next set of important attributes are quantities for available construction mass and different food types to digest (e.g. material harvested from plants or dead cells). When food matter enters a cell it is not automatically available for use as construction mass. It requires digestion to yield useful material and energy, and the rate at which this happens is controllable with a `digestion rate' variable. Finally, there are attributes for the quantities of available complex molecules and energy.

\begin{figure*}
\includegraphics[width=\linewidth]{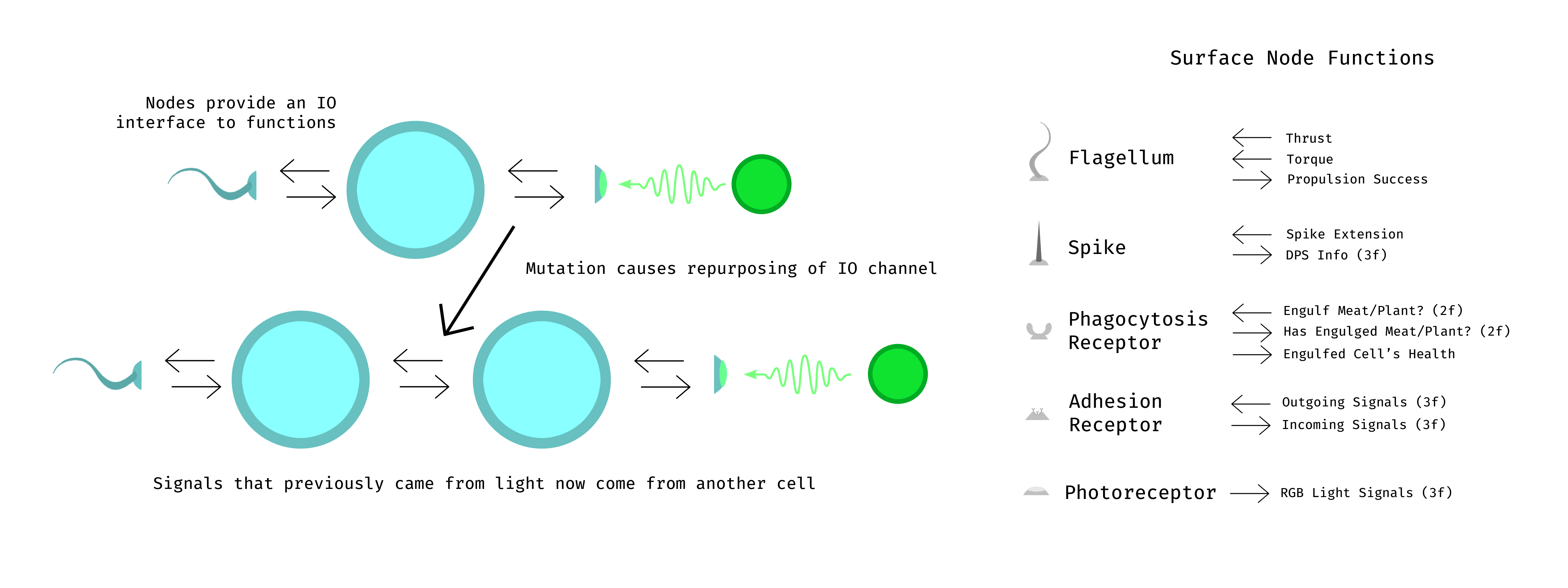}
\vspace{-1cm}
\caption{
    Overview of the surface nodes system. On the right there is a list of the possible node functional attachments, along with their respective graphical representations and IO semantics. The leftwards arrows represent signals going from the GRN into the attachment and the rightward arrows the reverse. Some IO channels are grouped together, for instance, ``DPS Info (3f)'' means three floating point numbers are returned that encode the DPS (Damage Per Second) information. Additionally, there are circumstances where some channels are not used (e.g. all of the input channels for photoreceptors).
    The left-hand diagram illustrates the interplay between the GRN and the modular surface nodes, using the hypothetical example found below.
} \label{fig:surface-node-system}
\end{figure*}

\textbf{Cell Types.} There are three types of cells in the simulation; `plant' cells, `meat' cells, and what we will call `protozoan' cells. The term protozoan is not derived from an intention to model any specific real-life species, but rather to highlight that this cell category in the simulation has access to typically protozoan-like functions such as motility, light sensitivity, and phagocytosis (engulfing other cells). The plant cells on the other hand, `photosynthesize', meaning that they spontaneously generate resources (mass and energy). Finally, meat cells are left behind when a protozoan cell dies, leaving fractions of the available energy, construction mass, and complex molecules in the meat cells.

\textbf{Energy and Resource Flows.} The primary sources of mass and energy in simulation are the plant cells. Plants are ingested by protozoan cells and converted into the stores of available energy and construction mass. Next there are a number of different directions these resources can flow; energy can be converted into action (e.g. in the form of movement). Mass and energy can be used to increase the cell's supply of complex molecules to be used for later construction projects. Such projects themselves will also require further construction mass and energy on top of the initial investment into the complex molecules. As mentioned before, upon a protozoan's death its supply of resources is distributed to meat cells spawned in its wake. This includes energy storage, construction mass, and complex molecules. The resources can then be reclaimed by other protozoans that ingest the meat. Meat is denser in energy than plant cells, and they present the potential to skip producing any stored complex molecules.

\textbf{Cell Growth.} Growth is the conversion of mass and energy into increasing the radius of the cell. This is mediated by a growth rate variable that is determined through genes (or when gene regulation is present, can also be controlled via environmental signals).

\textbf{Feeding.} The simulation provides two methods of feeding directly inspired by real protozoa; phagocytosis and osmotrophy. Phagocytosis is the process by which one cell engulfs another, drawing it into its interior and digesting it to extract resources. In the simulation, protozoa commence phagocytosis, killing the victim cell, removing it from the regular physics collision detection, and exterting a force on the prey towards the centre of the protozoan. The prey is then slowly killed (reducing their size) and material is extracted for processing in the protozoa. Multiple cells can be engulfed at once, but the total area of the prey cells cannot exceed 80\% that of the protozoan.

The second form of feeding is via osmotrophy from the \emph{chemical solution} that cells swim in (particularly visible in Figure \ref{subfig-1:chemical-soln-cells-feeding}). This aspect of the simulation can be seen in the graphical interface as the glowing effect surrounding plant or meat cells. At each moment in time, these cells deposit an amount of their mass into the environment. This is implemented via a $1024\times 1024$ RGB texture that overlays the  environment (excluding the void). Depositing onto this solution means drawing a filled circle that traces the cell with its colour onto the texture. The cell then has its colour linearly interpolated towards a desaturated greyer colour. The chemicals texture is then blurred with a $3\times 3$ box blur convolution to simulate diffusion.

When protozoa travel through these coloured regions the depositing process happens in reverse. For each pixel in the texture below the protozoa, the colour is extracted and food is added to the protozoa according to the region in colour space that the pixel occupies. Regions where the green component is greater than 0.5 and 1.5 greater than both the blue and red components could only be produced by plants, and so plant food is added proportionally. Likewise for meat and the red component.

Phagocytosis is only possible for protozoa that explicitly use resources to develop specialised receptors (see the next section), whereas osmotrophy happens by default for all protozoa. However, phagocytosis is much more lucrative as the transfer of resources through the chemical solution yields a very low extraction.

\subsection{Surface Nodes}
Each protozoan maintains an ordered list of \emph{surface node} objects that serve as sites for the development of specialised functions, referred to as \emph{node functional attachments}, such as flagella or photoreceptors. Each node is placed at a position on the cell's surface specified by a given angle. A node is an input/output (IO) channel between the internal control/gene regulatory system of the cell and the attachment. Before discussing the IO system in more detail, the following is a brief overview of each possible node attachment.

\begin{itemize}

\item \textbf{Flagella.} In real life, flagella are thin hair-like protrusions from a cell's surface that produce thrust and torque. Cells in the simulation can produce movement without the need for flagella, but a flagellum provides five times as much movement as the default. The thrust is generated through the vector from the node's position to the centre of the cell.

\item \textbf{Spikes.} Spikes are rigid protrusions that damage other cells when they come into contact. The damage induced is proportional to the depth the spike penetrates. The spike can be disabled by being retracted.

\item \textbf{Photoreceptors.} Photoreceptors produce signals derived from the red/green/blue (RGB) colour values of neighbouring obstacles. A number of rays are cast radially from the node's position and the resultant activations are the weighted average colours of the closest intersecting objects (where weights fall off quadratically with distance).

\item \textbf{Phagocytosis receptors.} These receptors facilitate feeding by engulfing cells. When a plant or meat cell comes into contact with a protozoan, if it is close enough to a phagocytosis receptor and the protozoan has enough capacity, the cell will be engulfed. The attachment takes two inputs corresponding to whether or not to engulf cells of a given type, and provides three outputs that encode the type of the most recently engulfed cell (that is still being absorbed), and that same cell's current health value.

\item \textbf{Adhesion receptors.} An adhesion receptor manages and creates bindings to another cell. In order for a binding to form, both cells need to have adhesion receptors. When they come into contact, the receptors find each other and form a joint anchored by the closest points on each cell to their partner. Two cells can only be bound by one adhesion receptor, and if there are multiple possible receptors that could produce the binding the ties are broken by order in the surface nodes list. Crucially this is stable across generations (unless explicitly mutated).

\end{itemize}

Regardless of which attachment is present (if any), four numbers go into the node: three of those modulate the function of the attachment, and the forth determines which function should be developed. We will refer to those first three variables as the control parameters, and the last as the \emph{construction signature} for the node. Similarly for the other direction; three numbers are returned from attachment and fed back into the control systems. These are called the sensor parameters. So for example, if a flagellum attachment is present on a node, the 1st of the three control parameters determine how much thrust it will attempt to generate, the 2nd determines the torque, and the 3rd is simply unused in this case. The flagellum then returns one number: the `propulsion success', and the other 2 are unused. As generating thrust or torque requires energy, a given input may be unachievable. If this is the case, whatever energy is available is used to generate the movement, and the propulsion success is the ratio between what was asked for and what was produced. In Figure \ref{fig:surface-node-system} the full set of attachment types and IO semantics can be found.

Figure \ref{fig:surface-node-system} also shows a diagram outlining a hypothetical example that highlights the relationship between the surface nodes system and the gene regulation. The illustration has two vertical layers. The top represents an initial set-up where a protozoan has a photoreceptor and a flagellum. Suppose that the internal control system has evolved to respond to green light and move the flagellum. The fact that surface nodes can be replaced with different functions means that these control systems can be repurposed, for instance by replacing the photoreceptor with a binding node and thus allowing a connecting cell to control the flagellum by sending replacing the green light signals, as shown in the bottom layer of the diagram.

Such reuse of pre-evolved systems is understood to be an important part of the evolution of complex biology, and is what Neil Shubin calls \textit{revolutionary repurposing} \citep{shubin_revolutionary_2020}. Critically for this to occur there is the need for gene regulation; if both cells are to have the same genes, they must have expression pathways that condition one cell to have a flagellum and the other a photoreceptor in the presence of the binding. In the later sections we will see an evolved example of such regulated node function growth in a run of the simulation.

\subsection{Fuzzy Lock-and-Key Algorithm}

At birth, the set of nodes is defined, but the attachments are not. Attachments are produced according to two factors: the node's \emph{construction signature} (the lock) and the cell's store of available \emph{complex molecules} (the keys). These are brought together to form the proposed \emph{fuzzy lock-and-key} procedure. The idea is to look at each molecule in the cell's storage and compute a `matching distance` between the construction signature $s_c$ and and the molecule's signature $s_m$,  $d_{matching} = D_{cycle}(s_c, s_m)$, where:
\begin{align}
D_{cycle}(s_1, s_2) &= \text{min}(s_{max} - s_{min},~1-s_{max}+s_{min}) \\
\text{where}~s_{max}&=\text{max}(s_1, s_2),\quad s_{min} =\text{min}(s_1,s_2)
\end{align}

Next, using a \emph{critical threshold} $d_{critical}$, we construct a \emph{matching coefficient} between the construction and available molecule signatures:
\begin{align}
k_{matching} = \begin{cases}
0 & \text{if}~d_{matching} \geq d_{critical} \\
\frac{d_{critical} - d_{matching}}{d_{matching}} & \text{otherwise}
\end{cases}
\end{align}

So when the matching distance is zero, the matching coefficient is one, when it is greater than or equal to the critical distance, it is zero, and the values in between are given by a linear interpolation.

This quantity can be interpreted as `the degree to which a given key opens a given lock', the next step is to define what is being unlocked. More specifically, how this ends in the node constructing a particular functional attachment. This is done by defining a \emph{functional potency} of the molecule with respect to each candidate attachment.

The first step in this is placing each of the attachment types on the unit interval at \emph{function points}. The types are ordered and then equally spaced: for $T$ types, type $i$ sits at function point $f_i = i/T$. The following ordering was used in the simulation runs presented in this article: 1) Flagellum, 2) Spike, 3) Phagoreceptor, 4) Photoreceptor, 5) Adhesion receptor. Then, when evaluating the functional potency for a given molecule we divide the distance to the closest function point $f_{closest}$ by $1/2T$, i.e. the distance at which the current function point would no longer be the closest. This distance can also serve as our critical matching distance for the fuzzy lock-and-key procedure. The functional potency is therefore:
\begin{align}
k_{func} &= 2T|s_m - f_{closest}|\\
\text{where}~f_{closest} &= \text{argmin}_{f_i} \{|s_m - f_i|~:~0 \leq i < T\}
\end{align}

Finally, the product $k_{func} k_{matching}$ of these two coefficients dictates how much the cell will attempt to progress the construction project for the attachment type at the closest function point. This all will only create an attempt because construction projects still require that given quantities of energy, mass, and complex molecules be present.

\textbf{Construction Projects.} Construction projects require four ingredients: mass, energy, complex molecules, and time. The time requirement dictates how much of the other three resources need to be contributed to the project over a given time step. If less that the required resources is available then progression can still be made, but at a slower rate proportional to ratio between what is available and what is required. For the complex molecule requirement, a molecule that does not exactly match can still be used to make progress, but this also incurs a penalty to the amount of progress made. 

\subsection{Reproduction and Evolution}

\begin{figure*}[!ht]
    \centering
    \subfloat[\label{subfig-1:multicellular-structures-size}]{%
          \includegraphics[width=.48\linewidth]{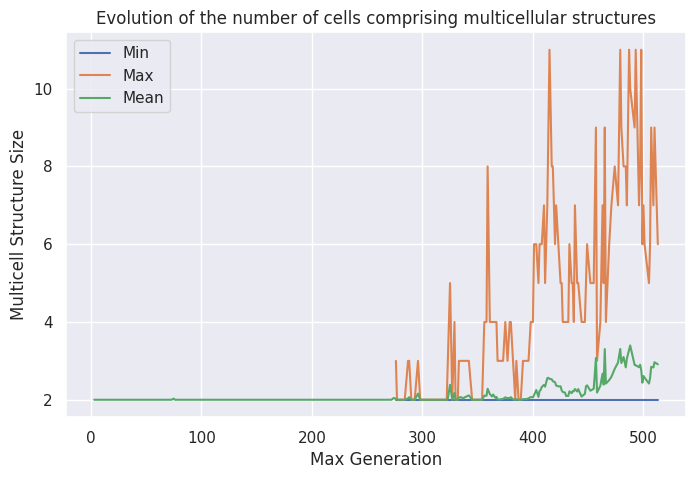}
    }
    \hspace{.2cm}
    \subfloat[\label{subfig-2:node-freq-changes}]{%
          \includegraphics[width=.48\linewidth]{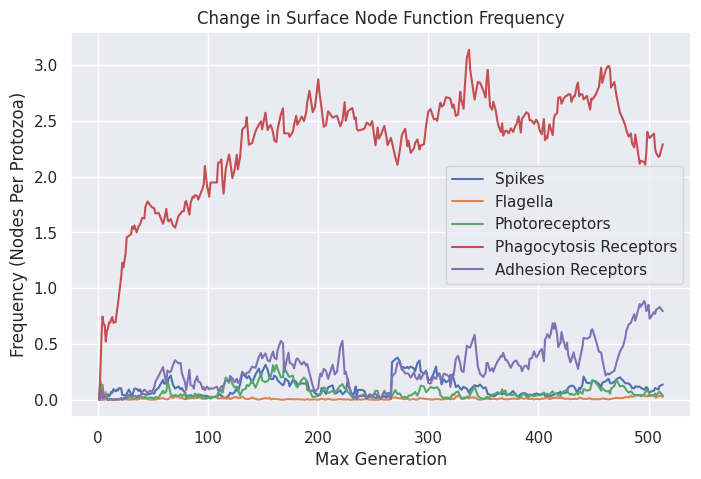}
    }
    \caption{Change in the sizes of multicellular structures (left) and changes in node type frequencies (right). The x-axis is the maximum of the generation across the protozoa alive at a given moment.}\label{fig:changes-nodes}
\end{figure*}

\textbf{Gene Regulatory Networks (GRNs).} Each protozoan cell in the simulation has a \textit{gene expression function} that handles input and output between the GRN and the cells attributes, sensors, and and control parameters. The GRN is modelled as a \emph{Recurrent Neural Network}, meaning that at any point in time each neuron node is maintaining a current state value. The whole network can then be `ticked' by synchronously computing each neuron's next state as a linear combination of its inputs passed through an activation function. The gene expression function is ticked at a regular interval. Before ticking the GRN, the various input values are loaded into the sensor neurons, and then after the output neuron states are unloaded onto the protozoan's attributes. In addition to the input neurons defined by the protozoan itself, there are two additional inputs; a \textit{bias} that is always equal to one and a \textit{random source} that samples a random value in the range of -1 to 1 on every tick. 

All inputs are provided in the range -1 to 1, and all outputs are mapped to a given output range by a cyclical linear remapping. This means that if the input exceeds either bound, it is brought back into the range using a modulo function. The use of this remapping is important for bounded traits such as the node construction signatures. The signatures need to be in the unit interval, but pre-activation values on the GRN outputs can be arbitrary large. Other mapping methods such as clamped linear remappings or sigmoid functions increase the density of the tails of the output range. For construction signatures this means that attachments on the ends of the interval (i.e. flagella and adhesion receptors) would be more likely when the weights of the GRN are randomly initialised.

\textbf{Initial Population.} From the perspective of the GRN, what we consider `control variables' (e.g. flagellum thrust) are indistinguishable from `traits' (e.g. node construction signatures). Likewise, `sensors' are indistinguishable from `gene regulators'. To the GRN these are just input and output nodes. Outside of explanatory aids, these differences are only relevant in the simulation at the point in which the initial populations are produced. Control variables are connected to any input neurons with 50\% randomly sampled connectivity and trait variables are only connected to the bias term. This is done to provide reproductive stability to the initial population as traits are reliably passed to the next generation (barring mutations). Otherwise traits such as construction signatures (and thus node attachments) would be randomly gene-regulated by default. If traits are not reliably inherited Darwinian selection could not occur.

\textbf{Reproduction.} Cells reproduce themselves asexually by cell division, which is triggered by reaching a size specified by the GRN and having adequate health to do so (greater than 0.15). This results in 2-6 child cells spawning in the parent's place, with larger cells being more likely to produce a larger number of offspring. At the moment of cell division there are two kinds of mutations that can happen. These relate to whether or not the mutable characteristic is part of the gene regulatory network (GRN) or not. There are only three unregulated (meaning not determined by outputs GRN) of traits: the number of nodes, the angles in which these nodes are placed on the cell, and the colour of the cell. The list of nodes can mutate in 3 ways, adding a node, deleting nodes, or changing the angle of a specific node. The colour changes by slightly increasing or decreasing the red, green, or blue components of the colour value. For the GRN itself, as it is represented by a neural network the mutation procedures outlined by the NEAT algorithm \citep{stanley_evolving_2002} are used.

\section{Results and Observations}

\begin{figure*}
    \centering
    \includegraphics[width=0.7\linewidth]{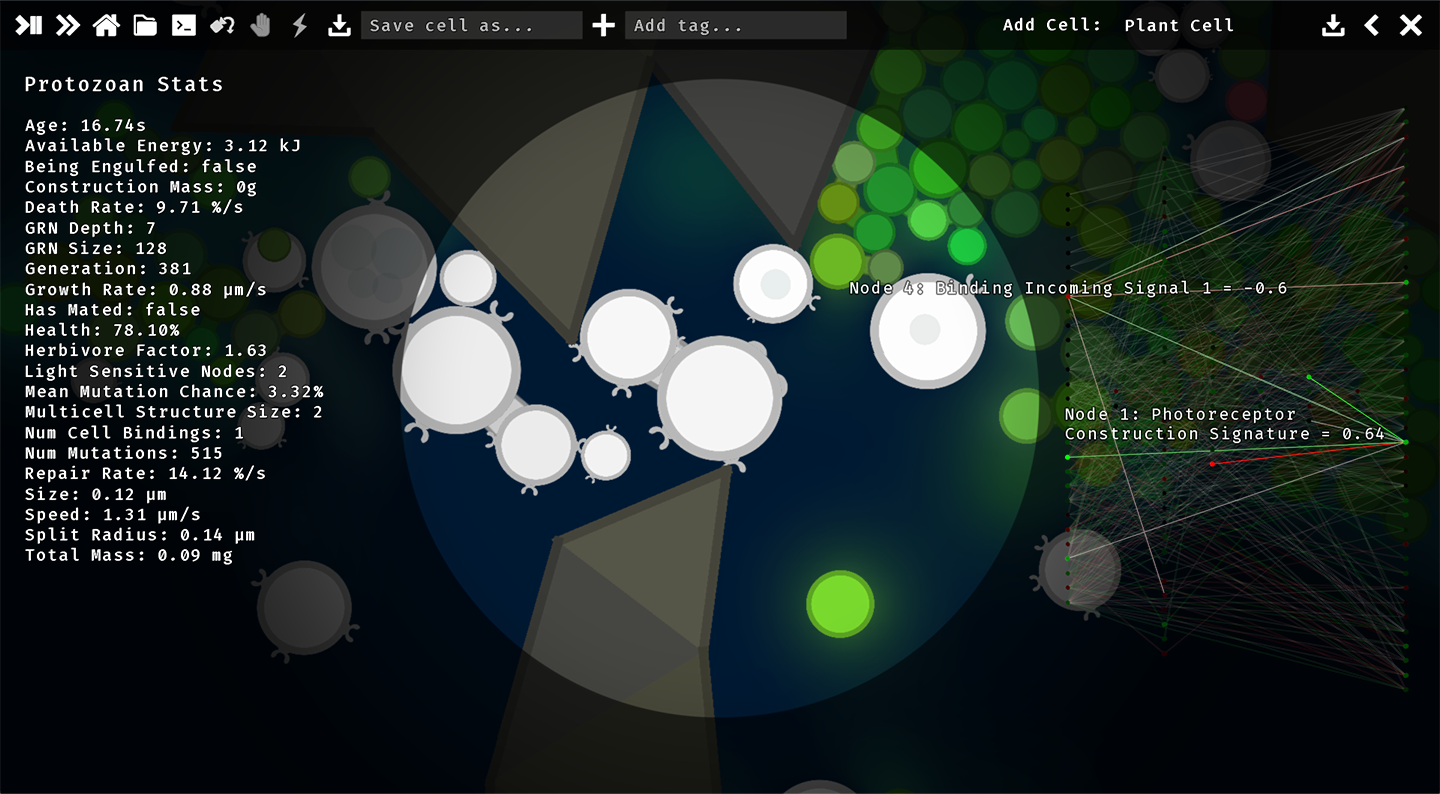}
    \caption{A view in the simulation of an evolved two-cell organism with a regulatory pathway between an incoming signal from another cell to the construction of a photoreceptor. Connections not involving the two IO nodes have reduced visibility.} \label{fig:binding-regulates-photo-construction}
\end{figure*}

The following sections look at results generated from a long run of the simulation over 3 days.

\subsection{Changes Over Generations}

Figure \ref{fig:changes-nodes} shows how protozoa traits changed over time. On the left we see the evolution of the size of multicellular structures. Note the minimum value is two, and the mean, maximum, and minimum lines refer to statistics taken from the sub-populations of cells that are involved in multicellular arrangements. For the first 250 generations the only multicellular structures that appeared were comprised of two cells, but after this point there was a steady increase in the mean and maximum sizes, with the largest structures emerging composed of 12 cells.

The right-hand plot (Figure \ref{subfig-2:node-freq-changes}) shows how the frequencies of the five node attachments changes over time. These frequencies are calculated as number of nodes present across all of the protozoa, divided by the total number of protozoa. As there are more nodes than there are protozoa, this results in the y-axis telling us how many nodes of a particular attachment the average protozoan has at any given point in time. 

At the beginning of the simulation we see a sharp rise in the number of phagocytosis nodes. This is expected as feeding by engulfing other cells provides so much more mass and energy than feeding by osmotrophy. Phagocytosis also cuts off the supply of nutrients in the chemical solution to nearby competitor protozoa. At around 200 generations into the run this ratio reaches around 2.5 phagocytosis nodes per protozoan, after which it fluctuates around roughly this value. Meanwhile, the other node types are coming and going, sometime reaching almost 50\% frequency, but it is not clear that these are cases of adaptive evolution instead of simply neutral evolution, or certain traits being taken along for the ride as they happen to be co-located in genomes with other genes that actually provide reproductive advantages. However, once adding more phagocytosis nodes reaches the point of diminishing returns (as having two nearby on the cell's surface serves no additional benefit), the emergence of larger multicellular begins within 100 generations. This begins with a sharp increase in the frequency of adhesion receptors and spikes at around generation 260. The high correlation of these two frequency curves at this point could indicate a single lineage becoming more prominent that had both node types. At around generation 330 these curves again diverge, with adhesion receptors steadily becoming more frequent and spikes disappearing. This perhaps shows a divergence of this lineage with a variance where the spike node being replaced with another adhesion node out-competes the original.

\subsection{Evolved Case Study: Cell Bindings Regulating the Construction of Photoreceptors.}

The section introducing the surface nodes system used an idealised hypothetical example of gene-regulated nodes to motivate the system's evolutionary importance. In Figure \ref{fig:binding-regulates-photo-construction} we see an multicellular organism that appeared during the run in which the presence of a cell binding regulates the development of photoreceptor nodes. In the figure we are looking at the UI tracking the right-hand cell of the organism, with the photoreceptors displayed as bumps on the top right of the cell. On the left we see a diagram of the cell's gene regulatory network, highlighting the connection between an incoming binding signal and the construction signature responsible for developing the photoreceptor on node 1.

Investigating the mechanisms of this arrangement we find the  regulatory pathway: first the adhesion receptors on each cell develop and they bind together. The binding incoming and outgoing signals are connected together, forming a feedback loop. With the other variables connected to these neurons being constant at birth\footnote{E.g. the health of the cell is connected but does not decrease rapidly enough to effect the dynamics of the circuit.}, this circuit quickly reaches a fixed-point attractor where the incoming and outgoing signals take stable but different values for each cell. The second incoming binding signal is then connected to the GRN output that controls the attachment construction signature for node 1. For one of the cells the signature corresponds to a phagocytosis node, and for the other it corresponds to a photoreceptor. For the cell with the photoreceptor on node 1, the incoming green light signal for this node feeds into the construction signature for node 2\footnote{Via an intermediate linear node connected to nothing else.}. This modification is what causes the second replacement of a phagocytosis receptor with another photoreceptor.

As with the observations made from the frequency plots in the previous section, it is not clear that this is a case of adaptive evolution. Indeed, this seems unlikely given that the photoreceptors were not seen to be playing an obvious role in increasing fitness. Additionally, there were not immediately any copies of structure the nearby, indicating that this morphology was perhaps relatively novel.
Nonetheless, this rudimentary `body plan' was seen to be relatively stable and the photoreceptors not actively harmful to fitness. 

After some time the organism seen in Figure \ref{fig:binding-regulates-photo-construction} reproduced itself when both cells split, resulting in four children that again paired up into two new child configurations. Of these two new pairs, one successfully recreated the parent's arrangement, but the other pair failed to form the binding in time and the nodes that would have developed into photoreceptors instead became phagocytosis nodes. Yet, coming back a few generations later, the photoreceptor configuration could still be seen copied across several two-cell organisms in the vicinity of the first sighting. This demonstrates that even if the photoreceptors were not adaptive at the initial sighting, there was the potential for evolution to build upon the newly developed morphology.

\section{Conclusions and Further Work}

The key components of the simulation are: 1) the resource flows of energy, mass, and complex molecules, 2) modular \textit{surface node} functions, and 3) the \textit{fuzzy lock-and-key} procedure for connecting the discrete cell function types to the gene regulatory network, allowing for cell specialisation. We have seen how the statistics of the simulation evolved and a case study showing a multicellular system in which cell specialisation occurred.

Further investigations will run the simulation for longer and under harsher survival conditions. Naturally, as a complex simulation there are many tunable parameters, for instance the maximum resource transfer rate across cells in multicellular structures, the energy densities of different foods, the amount of resources required for completing projects, ect. Adjusting these will certainly lead to different results and enhance our understanding of the dynamics discussed here. In the run of the simulation presented in the previous section it was seen that after 200 generations or so, reserves of complex molecules passed from parents to offspring resulted in most cells having an abundance of this resource. 

Another important observation from the runs done so far is that multicellular structures tend to appear first within the tighter areas of the environment between rock formations. This is probably because movement is harder for bound cells trying to go in opposing directions, which is a disadvantage in open spaces, but in enclosed areas colonies of cells can block off pathways and stop other cells that are not a part of the colony from accessing resources. This is indicative of another area further area of study; looking at the effects of more environmental diversity on evolving multicellular niches.

Finally, this work sits within a wider context of approaches for evolving simulating ecosystems and developing algorithms that leverage Darwinian principles for solving engineering problems. As every programmer appreciates, modularity is a key tool for developing complicated systems, and future work involving the ideas presented here will look at leveraging these algorithms towards other purposes for evolving more abstract modular systems.

\section{Acknowledgements}

Work is supported by the UKRI Centre for Doctoral Training in Safe and Trusted AI (EPSRC Project EP/S023356/1).

\footnotesize
\bibliographystyle{apalike}
\bibliography{references} 

\end{document}